%% file: main.tex
\documentclass[a4paper,conference]{IEEEtran}
\IEEEoverridecommandlockouts
\usepackage{cite}
\usepackage{subcaption}
\usepackage{amsmath,amssymb,amsfonts}
\usepackage{algorithmic}
\usepackage{graphicx}
\usepackage{textcomp}
\usepackage{algorithm}
\usepackage{xcolor}
\usepackage{placeins}
\usepackage[colorlinks=true,linkcolor=blue,citecolor=blue,urlcolor=blue]{hyperref}
\usepackage{float}
\usepackage{array}
\usepackage{enumitem}
\usepackage[left=1.57cm,right=1.57cm,top=1.9cm,bottom=2.54cm]{geometry}

\definecolor{deepskyblue}{rgb}{0.0, 0.25, 0.6}
\newcommand{\codekeyword}[1]{\textcolor{deepskyblue}{\textbf{#1}}}
\newcommand{\myfor}[1]{\STATE \textbf{\textcolor{violet}{for}} #1 \textbf{\textcolor{violet}{do}}}

\makeatletter 
\newcommand{\linebreakand}{%
  \end{@IEEEauthorhalign}
  \hfill\mbox{}\par
  \mbox{}\hfill\begin{@IEEEauthorhalign}
}
\makeatother 

\begin{document}

\title{
Evaluating Time Series Models for Urban Wastewater Management: Predictive Performance, Model Complexity and Resilience
\thanks{The authors gratefully acknowledge the financial support provided by the Federal Ministry for Economic Affairs and Climate Action of Germany for the RIWWER project (01MD22007H, 01MD22007C).}
}
\author{
  \IEEEauthorblockN{1\textsuperscript{st} Vipin Singh\textsuperscript{1, *}}
  \IEEEauthorblockA{vipin.singh@bht-berlin.de}
  \and
  \IEEEauthorblockN{2\textsuperscript{nd} Tianheng Ling\textsuperscript{2}}
  \IEEEauthorblockA{tianheng.ling@uni-due.de}
  \and
  \IEEEauthorblockN{3\textsuperscript{rd} Teodor Chiaburu\textsuperscript{1}}
  \IEEEauthorblockA{chiaburu.teodor@bht-berlin.de}
  \and
  \IEEEauthorblockN{4\textsuperscript{th} Felix Biessmann\textsuperscript{1, 3}}
  \IEEEauthorblockA{felix.biessmann@bht-berlin.de}
  \linebreakand
  \IEEEauthorblockA{
  \textsuperscript{1}Berlin University of Applied Sciences and Technology, Berlin, Germany\\
  \textsuperscript{2}Intelligent Embedded Systems Department, University of Duisburg-Essen, Duisburg, Germany\\
  \textsuperscript{3}Einstein Center Digital Future, Berlin, Germany\\
  \textsuperscript{*}Corresponding author.}
}

\maketitle

\begin{abstract}
Climate change increases the frequency of extreme rainfall, placing a significant strain on urban infrastructures, especially Combined Sewer Systems (CSS). Overflows from overburdened CSS release untreated wastewater into surface waters, posing environmental and public health risks. Although traditional physics-based models are effective, they are costly to maintain and difficult to adapt to evolving system dynamics. Machine Learning (ML) approaches offer cost-efficient alternatives with greater adaptability. 
To systematically assess the potential of ML for modeling urban infrastructure systems, we propose a protocol for evaluating Neural Network architectures for CSS time series forecasting with respect to predictive performance, model complexity, and robustness to perturbations. 
In addition, we assess model performance on peak events and critical fluctuations, as these are the key regimes for urban wastewater management.
To investigate the feasibility of lightweight models suitable for IoT deployment, we compare global models, which have access to all information, with local models, which rely solely on nearby sensor readings.
Additionally, to explore the security risks posed by network outages or adversarial attacks on urban infrastructure, we introduce error models that assess the resilience of models.
Our results demonstrate that while global models achieve higher predictive performance, local models provide sufficient resilience in decentralized scenarios, ensuring robust modeling of urban infrastructure. Furthermore, models with longer native forecast horizons exhibit greater robustness to data perturbations. These findings contribute to the development of interpretable and reliable ML solutions for sustainable urban wastewater management. 
The implementation is available in our GitHub repository\footnote{\url{https://github.com/calgo-lab/resilient-timeseries-evaluation}~\cite{githubRepo}}.
\end{abstract}


\section{Introduction}
\input{sections/introduction.tex}

\section{Related Work}
\input{sections/related_work.tex}

\section{Methodology}
\input{sections/methodology.tex}

\section{Robustness Analysis}
\input{sections/robustness_analysis.tex}

\section{Results and Discussion}
\input{sections/results.tex}

\FloatBarrier
\section{Conclusion and Future Work}
\input{sections/conclusion.tex}

\bibliographystyle{IEEEtran}
\bibliography{references}

\end{document}

%% file: sections/introduction.tex
Rapid urbanization is transforming landscapes globally, and more people are moving to cities than ever~\cite{ritchie2023urbanization}. This demographic surge intensifies the demands on aging urban infrastructure, particularly \emph{Combined Sewer Systems} (CSS), which were designed decades ago based on outdated climate assumptions~\cite{wilbanks2014climate}. CSS collects and transports stormwater and wastewater within a single pipe network, reducing construction costs and land use~\cite{cssVsSss}. Today however, these systems are experiencing unprecedented stress due to the increased frequency and intensity of extreme rainfall driven by climate change~\cite{bolan2023impacts}.

The expansion of sealed surfaces in urban environments increases stormwater runoff, increasing the likelihood of \emph{Combined Sewer Overflows} (CSOs), where untreated wastewater is discharged into natural water bodies~\cite{scalenghe2009sealing}. These overflows pose significant public health risks by contaminating water supplies and exposing cities to regulatory penalties and infrastructural degradation~\cite{sonone2020health}. Proactive management of CSS filling levels could mitigate these overflows, but existing approaches are based on physics-based hydraulic models. Although effective, these models require high-quality data for calibration, and are costly to implement, maintain, and adapt to changing environments~\cite{botturi2021combined}. Even after implementation, these systems require regular reconfiguration to maintain high performance~\cite{fach2009calibration}, making them impractical for real-time decision making.

Recent advances in sensor technology~\cite{cominola2015smartmeters} and data transmission infrastructure~\cite{khutsoane2017iot} have paved the way for \emph{Machine Learning} (ML) approaches in real-time urban water management~\cite{wang2013consequential}. In particular, time series forecasting models based on \emph{Neural Network} (NN) have shown the ability to capture complex temporal patterns, enabling accurate predictions of overflow events without requiring detailed physical system description. However, compared to traditional physics-based models, these NNs often suffer from limited interpretability, which can undermine stakeholder trust and practical adoption. Furthermore, their robustness to real-world data perturbations, such as sensor noise, missing data, or network outages, remains underexplored.

To address these challenges, this study conducts a comprehensive analysis of forecast consistency and robustness under data perturbations. The evaluation protocols are designed to reflect real-world challenges that impact urban wastewater management, including extreme weather events and potential adversarial attacks, which can lead to sensor or network outages. We investigate two modeling approaches: (1) global models, which leverage all available sensor data as covariates, and (2) local models, which rely solely on data from the target sensor, representing decentralized scenarios where global data access is restricted due to sensor failures or network disruptions. This comparative analysis assesses the resilience of each approach under varying conditions, ensuring reliable predictions in practical settings. To further study all relevant factors determining modeling choices, we develop realistic error models. These proposed methodologies are applied to multiyear data from a large urban wastewater network, providing empirical insights into model selection for effective and resilient urban wastewater management. 

%% file: sections/related_work.tex
This work builds on our previous empirical framework~\cite{singh2024datadrivenCSS}, which investigates the performance of global and local models for CSS forecasting (see \autoref{subsec:globalAndLocal}), focusing on predictive precision and computational complexity across random weight initializations. Our study extends this prior work in three key ways:
\begin{itemize}
    \item \textbf{Incorporation of Historical Weather Forecasts}: By integrating historical weather forecast data as input features, we assess the generalization capabilities of models in real-world deployment scenarios.
    \item \textbf{Introduction of Realistic Error Models}: We evaluate model resilience under sensor noise, malfunctions, and network outages, addressing critical challenges in urban wastewater monitoring.
    \item \textbf{Evaluation of Critical Events:}: We develop assessment protocols designed to measure model performance during peak overflow events, which are of primary concern in wastewater management.
\end{itemize}
Our main contribution lies in a comprehensive empirical evaluation based on novel assessment protocols that examine the feasibility of various ML approaches for urban wastewater management modeling. 
A key requirement of models in this context is interpretability, which remains a barrier to adopting NNs in critical infrastructure. Although recent work explores hybrid statistical-ML models to balance accuracy and explainability~\cite{chiaburu2024interpretable,januschowski2020mlVsStat}, these often require model-specific adaptations. Complementing existing work on interpretability in ML, we propose an evaluation protocol based on realistic errors, or \textit{perturbation}, of the data. Systematically corrupting input data (e.g., masking sensors, adding noise) helps to investigate model behavior with respect to their resilience. By introducing a perturbation-based evaluation of model robustness, we extend existing interpretability measures with a notion of resilience of model types.

%% file: sections/methodology.tex
This section details the model development process, including six NN architectures, the dataset used for training and testing, and the methodological framework adopted for hyperparameter optimization and model evaluation.

\subsection{Neural Network Architectures}
The six evaluated neural networks will be presented, highlighting their key characteristics and suitability for time series forecasting:

\begin{enumerate}
    \item \textbf{LSTM:} The \emph{Long Short-Term Memory} (LSTM) network uses memory cells with input, output, and forget gates to capture both short- and long-term dependencies in sequential data~\cite{hochreiter1998vanishing, hochreiter1997lstm, gers2000forget}. However, its one-step recursive prediction mechanism can lead to error accumulation in multi-step forecasting.
    
    \item \textbf{DeepAR:} Built on an autoregressive LSTM backbone, the DeepAR network employs \emph{Monte Carlo} sampling to produce probabilistic forecasts, quantifying predictive uncertainty~\cite{salinas2020deepar}. Nevertheless, its sequential prediction framework remains susceptible to error propagation across time steps.
    
    \item \textbf{N-HiTS:} The \emph{Neural Hierarchical Interpolation for Time Series Forecasting} (N-HiTS) network features a purely feed-forward architecture with hierarchical interpolation, decomposing time series into frequency components~\cite{challu2023nhits}. It eliminates recursive dependencies, improving computational efficiency and mitigating error accumulation over extended horizons.
    
    \item \textbf{Transformer:} The Transformer network leverages \emph{Multi-head Self-Attention} (MHSA) mechanisms and positional encoding to efficiently capture long-term dependencies in time series data and parallelize computation~\cite{vaswani2017attention, wen2022transformers} without sequential recurrence, albeit at a high computational cost.
    
    \item \textbf{TCN:} \emph{Temporal Convolutional Networks} (TCN) employ causal and dilated convolutions to model temporal dependencies efficiently~\cite{bai2018empirical}, although with limitations in capturing long-term patterns.
    
    \item \textbf{TFT:} The \emph{Temporal Fusion Transformer} (TFT) integrates MHSA mechanism, LSTM layers and gating mechanisms to capture both short- and long-term dynamics while enhancing interpretability through variable selection and quantile regression~\cite{lim2021temporal}.
\end{enumerate}

\subsection{Dataset}
The dataset used in this study is provided by \emph{Wirtschaftsbetriebe Duisburg}~\cite{wb_duisburg} and comprises three years (2021–2023) of hourly-resampled time series data from six locations in Duisburg's Vierlinden district\footnote{Data cannot be publicly released. Readers may contact the corresponding author for further details.}. It includes 35 sensor features capturing various aspects of the CSS operation, such as water levels, pump energy consumption, valve states, and rainfall measurements. The target variable for our analysis are the filling levels of an overflow basin. Originally, sensor readings were event-driven, recorded at irregular intervals ranging from 1 second to 1 hour. To establish a uniform temporal resolution, the data were resampled to hourly intervals by computing the mean of values closest to each full-hour mark. Missing rainfall measurements were imputed using data from the nearest \emph{Deutsche Wetterdienst} station. 
Missing values in other features were interpolated using linear interpolation, with indicator columns flagging imputed entries to maintain data integrity. In addition, historical rain forecasts from the \emph{Open-Meteo Weather} API~\cite{zippenfenig2023openmeteo} were incorporated as future covariates, enabling the models to integrate physics-based predictions into the data-driven learning approach.

\subsection{Global vs. Local Model Approach}
\label{subsec:globalAndLocal}
This study evaluates two forecasting approaches for urban wastewater management:

\begin{enumerate}
    \item \textbf{Global Model:} Incorporates all available sensor data (35 features) under normal conditions, utilizing historical rain forecasts from the Open-Meteo API~\cite{zippenfenig2023openmeteo} as future covariates. By leveraging system-wide data and meteorological context, it is supposed to achieve higher predictive accuracy.
    \item \textbf{Local Model:} Operates autonomously during sensor failures or network disruptions, relying only on historical data from a given sensor. For models requiring future covariates (e.g., TFT), a placeholder feature created by shifting the target variable forward by the forecast horizon, substitutes unavailable external data. This improves resilience for edge deployment in critical infrastructure scenarios~\cite{ling2024flowprecision}.
\end{enumerate}

The global model prioritizes precision through rich contextualization, while the local model ensures operational continuity in scenarios where centralized data access is compromised.

\subsection{Model Development Process}
\label{subsec:model_development}

To preserve temporal dependencies and seasonal patterns, the dataset was chronologically split into training (January 2021–July 2022), validation (August–December 2022), and test (January–December 2023) sets. In our study, we used an input sequence length of 72 hours, a forecast horizon of 12 hours, and a batch size of 256. Hyperparameter tuning was performed separately for the global and local models using \emph{Optuna}'s~\cite{optuna} \emph{Tree-structured Parzen Estimator}~\cite{watanabe2023tpeUnderstanding} in a two-phase search over 600 trials (500 for broad exploration followed by 100 refined trials). The best configurations (see our GitHub Repository) for each model type were then trained 100 times with different random seeds to assess the consistency of forecasts. Training used the \emph{Adam} optimizer with early stopping (patience: 10 epochs) to mitigate overfitting, using MSE as the loss function. The experiments were carried out on an NVIDIA A100 GPU to meet computational demands.

%% file: sections/robustness_analysis.tex
This section presents the robustness analysis conducted in this study, evaluating the performance of the proposed models under real-world data quality issues that may impact forecasting accuracy.

\subsection{Error types}
\label{subsec:error_types}
\begin{figure}[ht]
    \centering
    \begin{tabular}{@{}m{0.15\linewidth}@{~ }m{0.7\linewidth}@{}}
        & \hspace{2.2em} {Original} \hspace{4.55em} {Corrupted} \\ 
        {Outliers:} &
        \includegraphics[width=\linewidth]{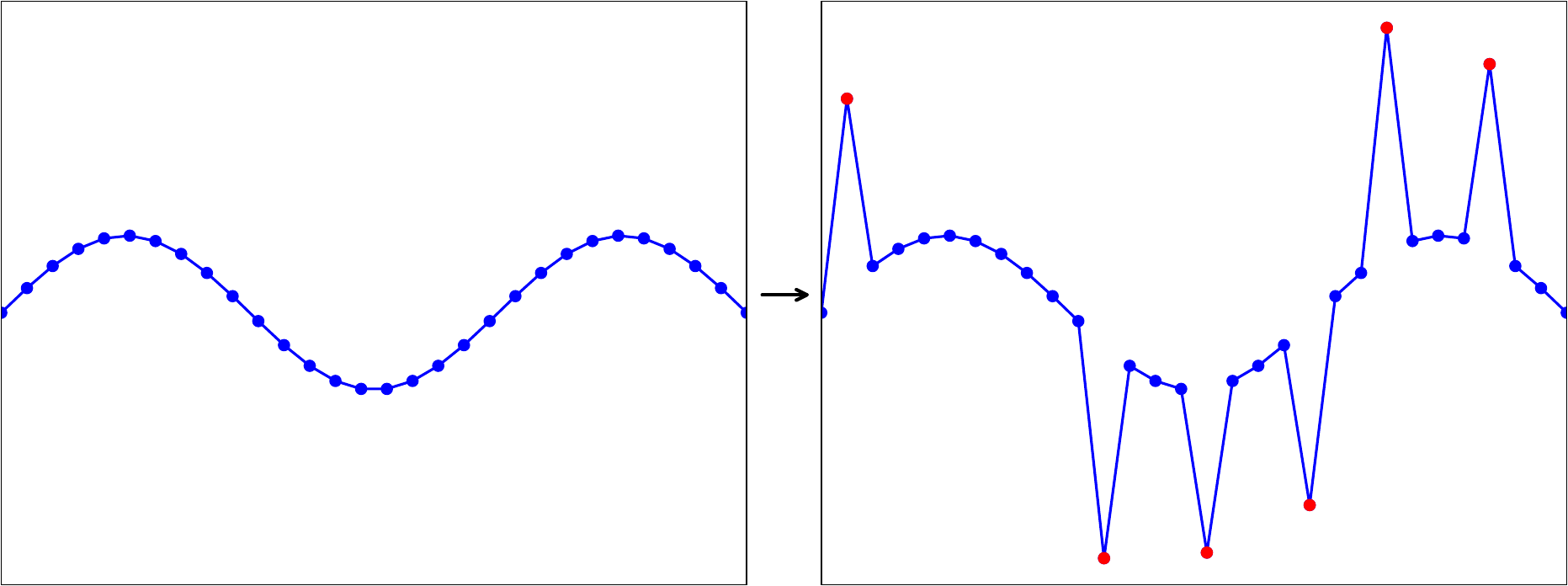} \\ 
        {Missing Values:} &
        \includegraphics[width=\linewidth]{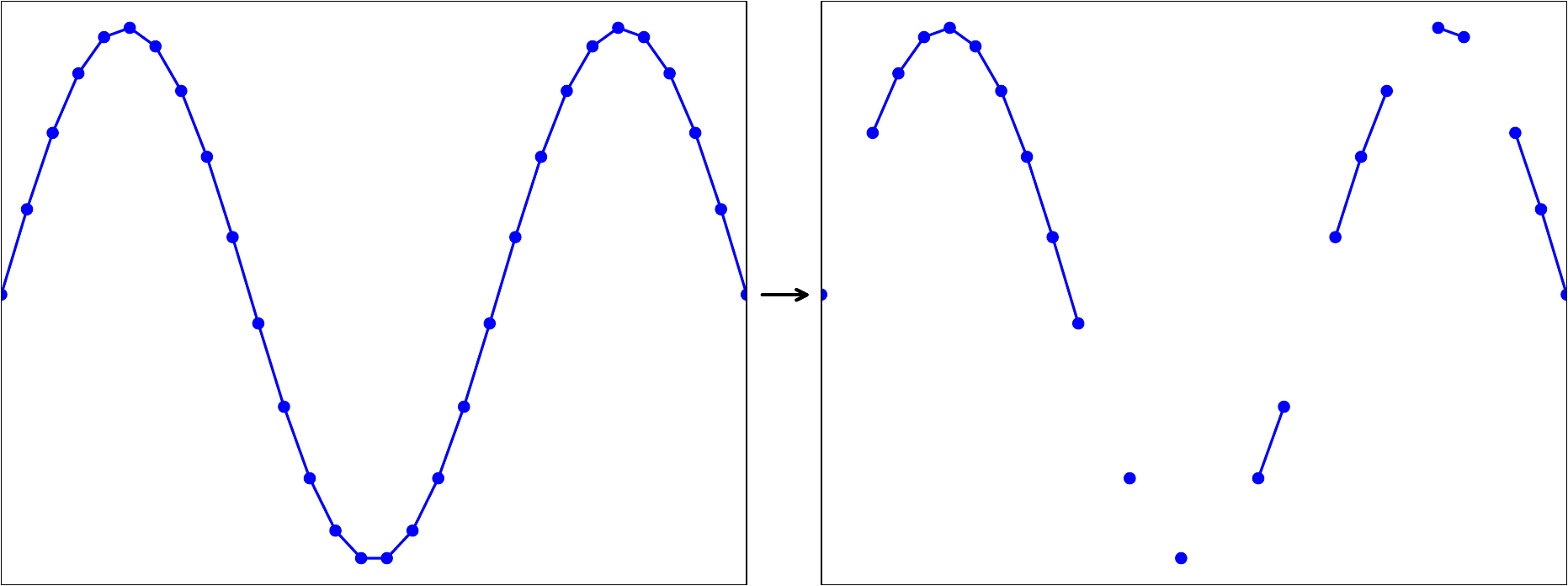} \\ 
        {Clipping:} &
        \includegraphics[width=\linewidth]{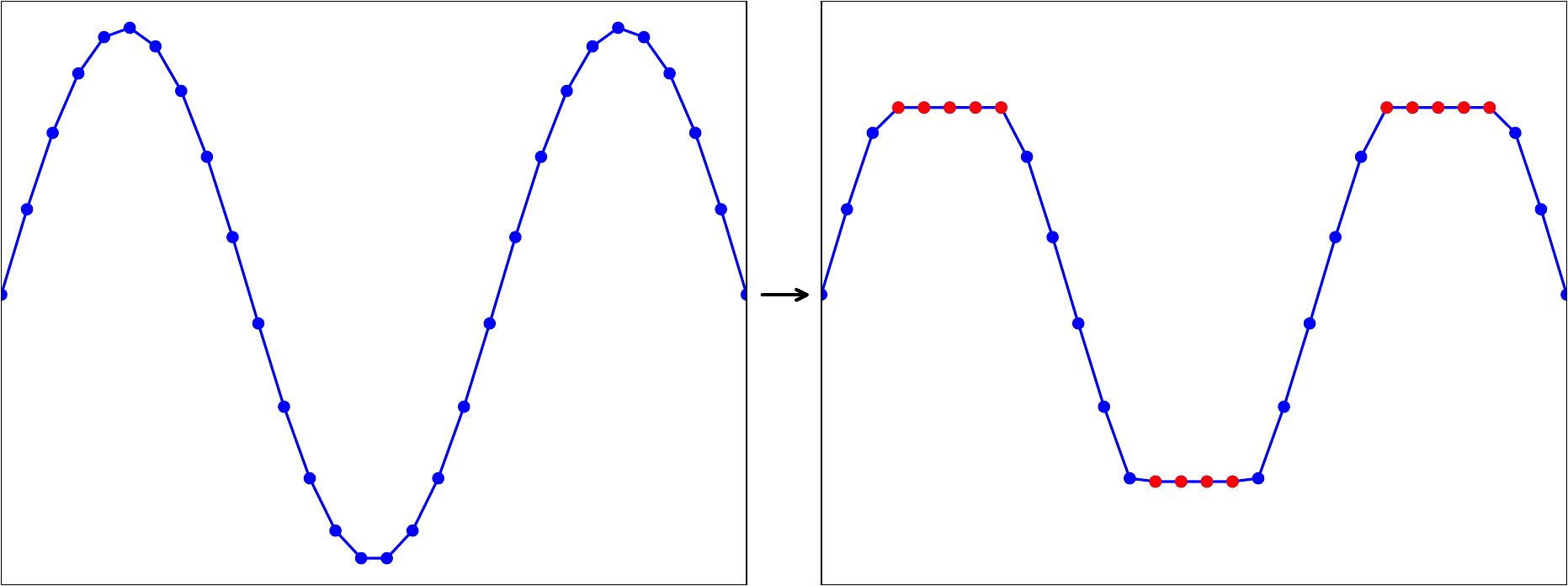} \\ 
    \end{tabular}%
    \caption{The three error types applied to dummy data.}
    \label{fig:errortype_comparison}
    \vspace{-15pt}
\end{figure}

To simulate common data perturbations encountered in sensor-based systems, we introduce three error types, visualized in \autoref{fig:errortype_comparison}.

\begin{enumerate}
    \item \textbf{Outliers} refer to data points that deviate significantly from the majority of observations, often caused by measurement errors or sensor miscalibration. They are identified using the \emph{Interquartile Range} (IQR) method, as defined in Equation \ref{eq:outliers}, where \(\text{IQR} = Q_3 - Q_1\).
    {
    \begin{equation}
        \begin{aligned}
            x &\notin [Q_1 - 1.5\,\text{IQR},\, Q_3 + 1.5\,\text{IQR}]
        \end{aligned}
    \label{eq:outliers}
    \end{equation}
    }
    To introduce controlled outlier perturbations, the values are modified as shown in Equation \ref{eq:outlier_modification}, where \(\mu\) denotes the mean of the data, and \(\mathcal{N}\) represents a \emph{Gaussian} distributed random number. The parameter \(\alpha\) controls the magnitude of the shift, while \(\beta\) determines the noise level. Note that when \(\alpha > 1\), the perturbed data points are guaranteed to be outliers.
    {
    \begin{equation}
    x_{\text{outlier}} =
        \begin{cases}
            x - \alpha \big(\mu - (Q_1 - 1.5\,\text{IQR})\big) \\
            \quad + \mathcal{N}(0, \beta\,\text{IQR}), & \text{if } x < \mu \\[0.5em]
            x + \alpha \big((Q_3 + 1.5\,\text{IQR}) - \mu\big) \\
            \quad + \mathcal{N}(0, \beta\,\text{IQR}), & \text{if } x > \mu
        \end{cases}
    \label{eq:outlier_modification}
    \end{equation}
    }
    
    \item \textbf{Missing Values} represented as \(x_{\text{missing}} \!=\! \text{NaN}\), simulate gaps in sensor readings, which commonly arise due to hardware failures, maintenance procedures, or communication issues.
    
    \item \textbf{Clipping} constrains data within predefined bounds to reflect sensor limitations. As shown in Equation \ref{eq:clipped}, \(Q_{\text{lower}}\) and \(Q_{\text{upper}}\) are the specified quantile bounds. This error type mimics scenarios where physical or design limits prevent sensor readings from exceeding certain values.
    {
    \begin{equation}
         x_{\text{clipped}} = 
        \begin{cases} 
        Q_{\text{lower}} & \text{if } x < Q_{\text{lower}} \\[1ex]
        x & \text{if } Q_{\text{lower}} \leq x \leq Q_{\text{upper}} \\[1ex]
        Q_{\text{upper}} & \text{if } x > Q_{\text{upper}}
        \end{cases}
    \label{eq:clipped}
    \end{equation}
    }
\end{enumerate}

\subsection{Error sampling}
Errors are injected using the \emph{tab\_err}\footnote{\url{https://github.com/calgo-lab/tab_err}~\cite{githubRepoTabErr}} error generation package~\cite{githubRepoTabErr}, which models realistic errors for tabular and time series data. The package supports patterns with complex dependencies between covariates and errors~\cite{jagerDataImputationData2024}.
In our study, the \emph{Errors-At-Random} scenario was employed, treating the timestamp as the dependent variable to ensure that errors appear in consecutive clusters, mimicking sensor downtimes or intermittent failures. The error rates are adjustable. However, for \emph{Value Clipping}, only values outside the specified bounds are modified, which can yield a lower effective error rate.


\subsection{Error Generation Setup}
\label{subsec:errorgen_setup}
The robustness evaluation follows a hierarchical structure represented by nested loops:
\begin{algorithm}
    \caption{Hierarchical Structure of Robustness Evaluation}
    \begin{algorithmic}[1]
        \myfor{each \codekeyword{Model Type}}
        \begin{ALC@g}
            \myfor{each \codekeyword{Model} $i$ in 100 Trials (\autoref{subsec:model_development})}
                \begin{ALC@g}
                \myfor{each \codekeyword{Feature}}
                    \begin{ALC@g}
                    \myfor{each \codekeyword{Error Type}}
                        \begin{ALC@g}
                        \myfor{each \codekeyword{Error Rate}}
                            \begin{ALC@g}
                            \STATE Evaluate robustness of model $i$ under given conditions
                            \end{ALC@g}
                        \end{ALC@g}
                    \end{ALC@g}
            \end{ALC@g}
        \end{ALC@g}
    \end{algorithmic}
\end{algorithm}

This ensures a systematic evaluation of the robustness of the models. For the error rates 10\%, 20\%, 30\%, 40\%, and 50\% were considered. Finally, these are the error type configurations, as defined in \autoref{subsec:error_types}:
\begin{itemize}
    \item Outliers: $\alpha = 1.1$, $\beta = 0.1$
    \item Missing Values: --
    \item Clipping: $q_{\text{lower}} = 0.2 \rightarrow Q_{0.2}$, $q_{\text{upper}} = 0.8 \rightarrow Q_{0.8}$
\end{itemize}

%% file: sections/results.tex
This section describes the results of the models and compares their performance in terms of predictive performance, robustness, and computational complexity.


\subsection{Predictive Performance}
\begin{figure}[h]
    \vspace{-3pt}
    \centering
    \includegraphics[width=0.9\linewidth]{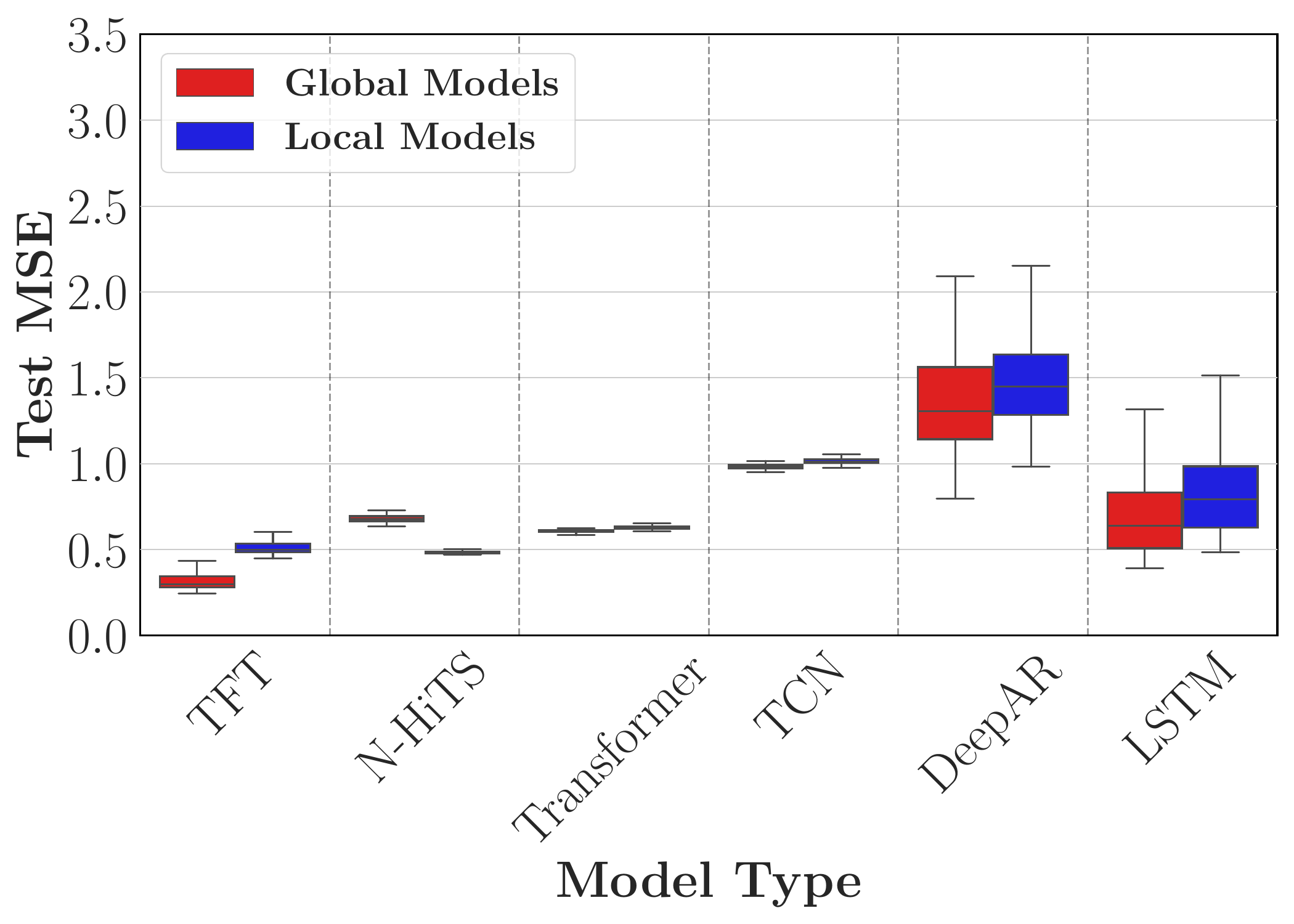}
    \caption{MSE for various NN architectures. Error bars represent variability across model initializations. Most models achieve similar performance, while LSTM and DeepAR show considerable variability across initializations. Global models often only slightly outperform local models.}
    \label{fig:mse_comparison}
\end{figure}

The predictive performance of the evaluated NN architectures, measured by \emph{Mean Squared Error} (MSE), is shown in \autoref{fig:mse_comparison}. To assess the consistency of the models, each architecture was trained 100 times with different random weight initializations, and the spread of MSE values across runs is reported.
\autoref{fig:mse_comparison} suggest that local models often achieve MSEs comparable to global models.
This trend can be seen across all model architectures evaluated. Upon closer investigation of the results, we found that there are long periods of time when there is little activity in the sewer system and sensor readings.

\begin{figure}[h]
    \vspace{-3pt}
    \centering
    \includegraphics[width=0.9\linewidth]{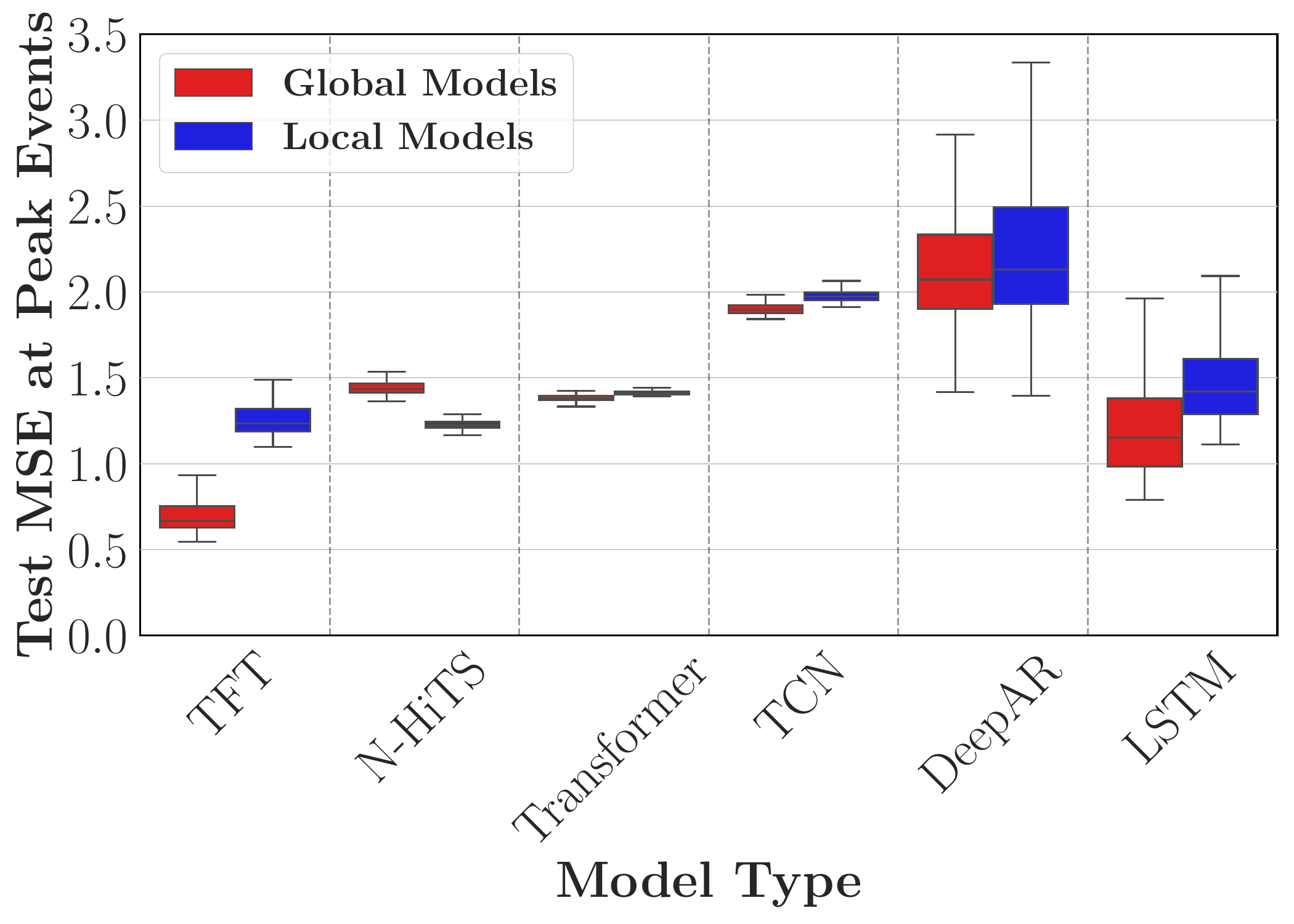}
    \caption{MSE during peak events shows a more pronounced difference between global and local models across NN architectures. Notably TFT performs much better at peak events than the other models, in particular the global TFT model.}
    \label{fig:mse_peak_events_comparison}
    \vspace{-2pt}
\end{figure}

To better investigate the predictive performance in regimes relevant for urban infrastructure, we conducted another evaluation in which we considered only peak events and critical fluctuations.
These events are identified by calculating the differences between consecutive values and selecting the highest 20\% of these differences. To smooth the time intervals, a rolling mean is applied before selection with a window size of 48, corresponding to 2 days. The MSEs of these events are shown in \autoref{fig:mse_peak_events_comparison}, suggesting that during peak events, global models perform better than local models, with the TFT model achieving the highest predictive precision.
%

\subsection{Robustness to Perturbations}
\begin{figure}[h]
    \vspace{-3pt}
    \centering
    \includegraphics[width=\linewidth]{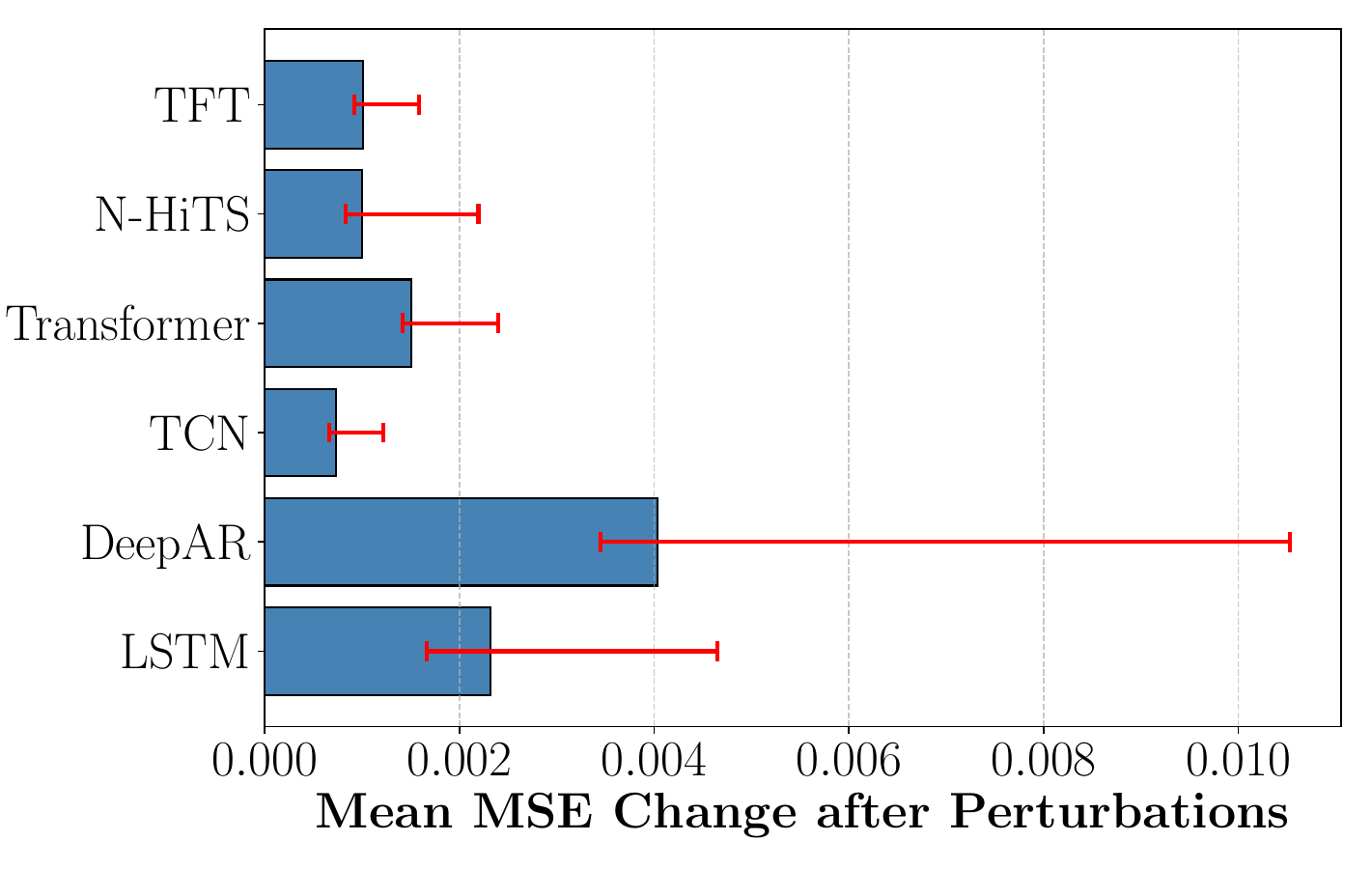}
    \caption{MSE increase between unperturbed and perturbed data for all models aggregated over error types, error rates, and features. The figure reveals that the DeepAR and LSTM models are most sensitive to the perturbations.}
    \label{fig:robustness_comparison}
    \vspace{-10pt}
\end{figure}

We investigated the robustness to realistic errors such as sensor noise, malfunctions, or network outages by applying outliers, missing values, and clipping as specified in \autoref{subsec:errorgen_setup}. \autoref{fig:robustness_comparison} shows all the MSE differences between unperturbed and perturbed data aggregated per model type.

The best performing model in terms of robustness is the TCN model, followed closely by the TFT and N-HiTS models.

%
\FloatBarrier
\subsection{Trade-Offs: Precision, Complexity and Robustness}

After individually evaluating the established metrics, the trade-offs between predictive performance, computational complexity, and robustness are analyzed in this subsection. 
The predictive performance is measured with the MSE, the computational complexity with the model size and inference time, and the robustness with the forecast consistency (IQR of the MSE) and the MSE increase due to perturbations.
To provide a more holistic view of the trade-offs, a \emph{Computation Complexity Index} (CCI) and a \emph{Robustness Index} (RI) are introduced:
{
\begin{equation}
    \text{CCI}_i = \frac{1}{2} \left( \frac{t_i}{\max(t)} + \frac{s_i}{\max(s)} \right)
    \label{eq:complexityIndex}
\end{equation}
}
with $t_i$ being the inference time, $s_i$ the model size of model $i$ and $\max(t)$ and $\max(s)$ the maximum inference time and model size among all models, respectively.
{
\begin{equation}
    \text{RI}_i = \frac{1}{3} \left( \frac{\text{iqr}_i}{\max(\text{iqr})} + \frac{\text{MSE}_{\text{pert},i}}{\max(\text{MSE}_{\text{pert}})} + \frac{\text{iqr}_{\text{pert},i}}{\max(\text{iqr}_{\text{pert}})} \right)
    \label{eq:robustnessIndex}
\end{equation}
}
%
with $\text{iqr}_i$ being the IQR of the MSE, $\text{MSE}_{\text{pert},i}$ the mean absolute MSE increase due to perturbations and $\text{iqr}_{\text{pert},i}$ the IQR of the absolute MSE increase due to perturbations of model $i$. The maximum values are calculated analogously to the complexity index.

\begin{figure}[!htb]
    \centering
    \begin{subfigure}{0.9\linewidth}
        \centering
        \includegraphics[width=\linewidth]{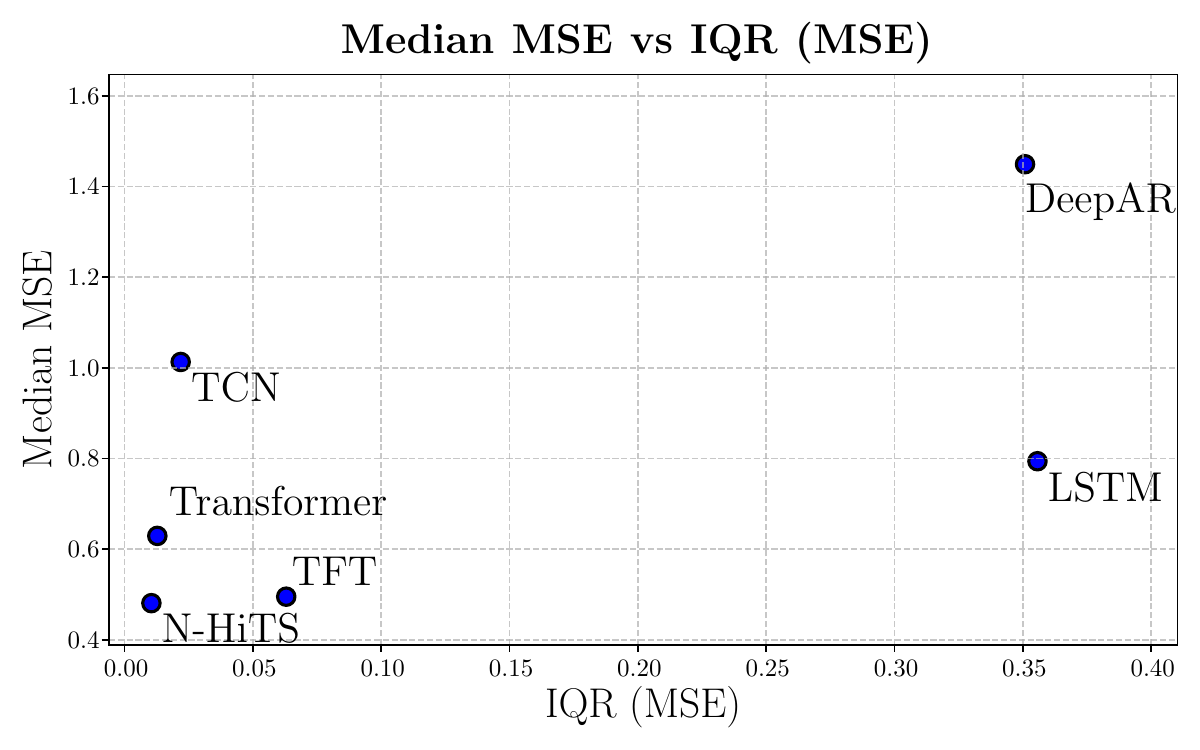}
    \end{subfigure}
    
    \vspace{-0.5mm} 

    \begin{subfigure}{0.9\linewidth}
        \centering
        \includegraphics[width=\linewidth]{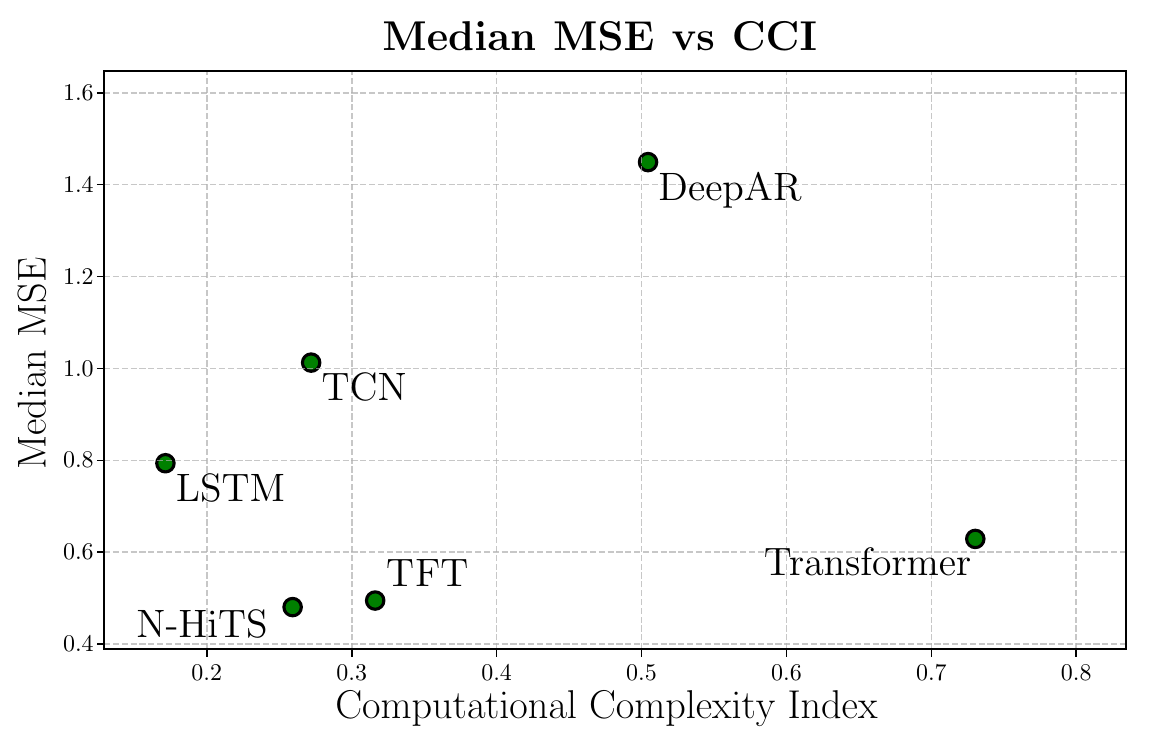}
    \end{subfigure}
    
    \caption{Trade-off analysis for \textbf{local models}: MSE vs. IQR (top) and MSE vs. CCI (bottom). N-HiTS and TFT show the best trade-off between predictive performance, computational complexity, and robustness.}
    \label{fig:tradeoff-local}
    \vspace{-10pt}
\end{figure}

\begin{figure}[!htb]
    \centering
    \begin{subfigure}{0.9\linewidth}
        \centering
        \includegraphics[width=\linewidth]{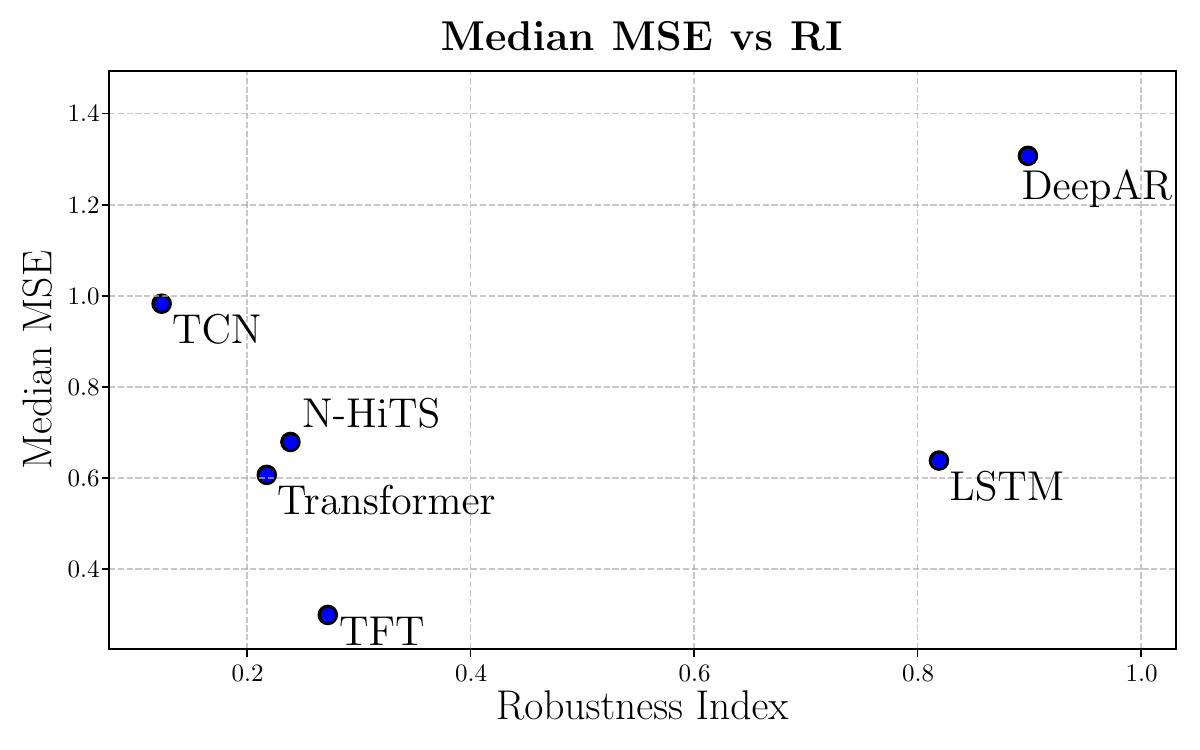}
    \end{subfigure}
    
    \vspace{-0.5mm} 

    \begin{subfigure}{0.9\linewidth}
        \centering
        \includegraphics[width=\linewidth]{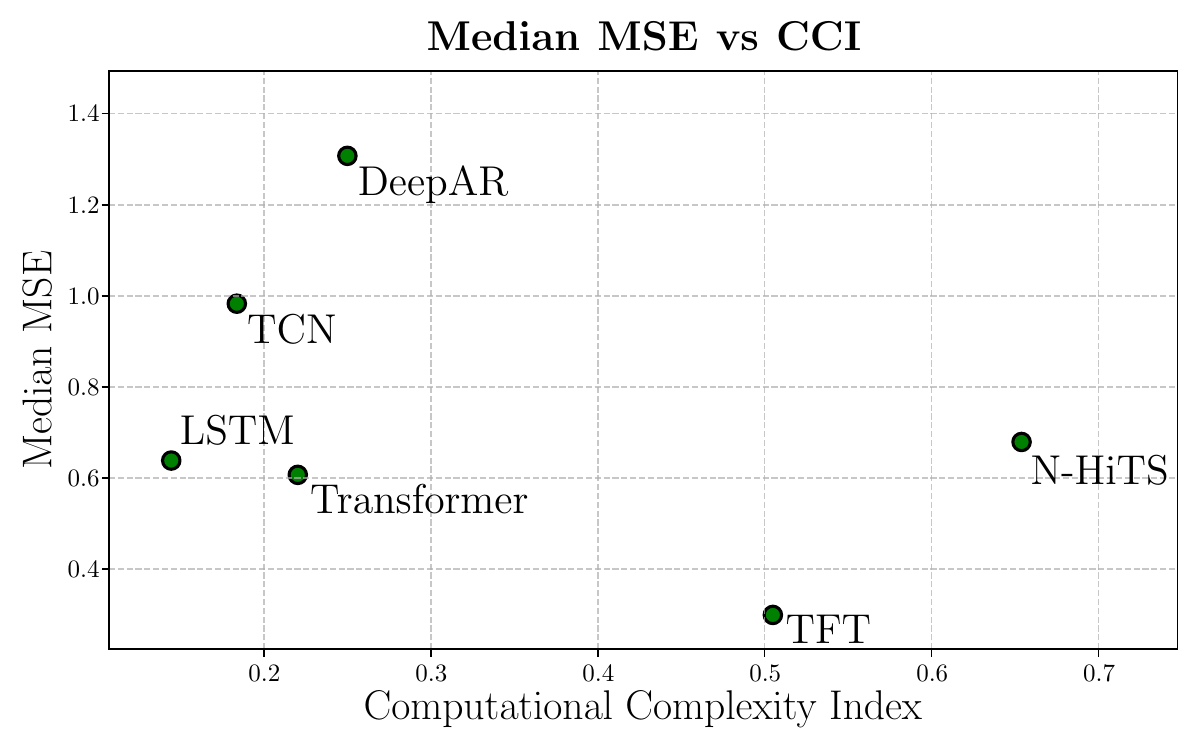}
    \end{subfigure}
    
    \caption{Trade-off analysis for \textbf{global models}: MSE vs. RI (top) and MSE vs. CCI (bottom). Transformer shows the best balance between the metrics, but has a poor predictive performance. In contrast TFT has a poor complexity, but a good predictive performance and robustness.}
    \label{fig:tradeoff-global}
\end{figure}

Due to the local model approach already simulating the extreme case of a total network outage, they have 100\% missingness in exogenous data. This effectively means that the perturbation analysis was not conducted for the local models, as they are inherently subject to the most severe perturbation. To assess their robustness, only the IQR of the MSE, representing the forecast consistency, is considered. 
The relationships between predictive performance, computational cost, and robustness are visualized in \autoref{fig:tradeoff-local} for local models and in \autoref{fig:tradeoff-global} for global models.
%
%

\autoref{fig:tradeoff-local} shows that the local N-HiTS and TFT models both exhibit high predictive performance. The N-HiTS model achieves this predictive performance at a computational complexity lower than that of TFT. The LSTM and DeepAR models perform poorly in terms of robustness, while DeepAR performs worst in predictive precision. The Transformer model has the worst complexity among the models.
%
%

For the global models there is no clear winner in the trade-off analysis, according to \autoref{fig:tradeoff-global}. The Transformer model has the best balance between the metrics, but has a poor predictive performance. The TFT model has a poor complexity, but a good predictive performance and robustness. The DeepAR and LSTM models show a poor robustness among the global models as well.
Overall the TFT model seems to achieve the best trade-off, it achieves superior predictive performance and very good robustness. However its complexity make it more difficult to deploy on edge devices.

%% file: sections/conclusion.tex
This study empirically evaluated six NN architectures for time series forecasting in CSS, focusing on predictive perfomance, computational complexity, and robustness under data perturbations. In a comprehensive empirical evaluation on a real-world dataset, we find that global models leveraging system-wide sensor data achieved superior accuracy, particularly the TFT, which excelled in predicting critical peak overflow events. Local models, relying solely on single-sensor data, demonstrated resilience in decentralized scenarios, with N-HiTS and TFT offering the best balance of performance and robustness. Multi-step forecasting models (e.g., N-HiTS, TCN, TFT, Transformer) proved significantly more robust to data perturbations, such as outliers, missing values, and clipping, than single-step autoregressive models (e.g., LSTM, DeepAR), highlighting the impact of error propagation in recursive architectures. Another key finding is that evaluation protocols that account for peak events may yield different results than when considering all data.

These findings confirm the potential of modern NNs to enhance CSS management in real-time. By integrating global models during normal operations and deploying local models as fail-safes during sensor failures, cities can mitigate overflow risks while maintaining operational continuity. The perturbation analysis further provides stakeholders with actionable insights into model reliability under real-world data quality challenges.

Future research should apply this framework to data from other cities to generalize the findings. To advance toward practical IoT deployment, model weight quantization techniques should be investigated to further reduce computational complexity, and the feasibility of models on embedded hardware should be systematically assessed. Additionally, evaluating grouped feature perturbations could offer deeper insight into how correlated errors affect model performance. Finally, a direct comparison with physics-based models remains an important but challenging next step.